\documentclass{article}

\usepackage{arxiv}

\usepackage[utf8]{inputenc} 
\usepackage[T1]{fontenc}    
\usepackage{hyperref}       
\usepackage{url}            
\usepackage{booktabs}       
\usepackage{amsfonts}       
\usepackage{nicefrac}       
\usepackage{microtype}      
\usepackage{lipsum}
\usepackage{graphicx}
\usepackage{xcolor}
\usepackage{multirow}
\graphicspath{ {./images/} }
\usepackage{tikz}

\title{Contactless Cardiac Pulse Monitoring Using Event Cameras}

\author{%
Mohamed Moustafa$^{1, 2}$ \quad Joseph Lemley$^{3}$ \quad Peter Corcoran$^1$\\
$^1$C3I Imaging Lab, School of Engineering, University of Galway, H91 TK33 Ireland \\
$^2$Autosense Team, FotoNation-Tobii, Galway, H91 V0TX Ireland \\
$^3$School of Computer Science, University of Galway, Galway, H91 TK33 Ireland\\
\texttt{\{m.moustafa1,joseph.lemley,peter.corcoran\}@universityofgalway.ie}
}

\begin{document}
\maketitle
\begin{abstract}
Time event cameras are a novel technology for recording scene information at extremely low latency and with low power consumption. Event cameras output a stream of events that encapsulate pixel-level light intensity changes within the scene, capturing information with a higher dynamic range and temporal resolution than traditional cameras. This study investigates the contact-free reconstruction of an individual's cardiac pulse signal from time event recording of their face using a supervised convolutional neural network (CNN) model. An end-to-end model is trained to extract the cardiac signal from a two-dimensional representation of the event stream, with model performance evaluated based on the accuracy of the calculated heart rate. The experimental results confirm that physiological cardiac information in the facial region is effectively preserved within the event stream, showcasing the potential of this novel sensor for remote heart rate monitoring. The model trained on event frames achieves a root mean square error (RMSE) of 3.32 beats per minute (bpm) compared to the RMSE of 2.92 bpm achieved by the baseline model trained on standard camera frames. Furthermore, models trained on event frames generated at 60 and 120 FPS outperformed the 30 FPS standard camera results, achieving an RMSE of 2.54 and 2.13 bpm, respectively.
\end{abstract}


\section{Introduction}
\label{sec:introduction} 
Computer vision, the study of computer comprehension and response to visual information, has been one of the main ways that modern technology has impacted people's personal and professional lives. A key contributing factor to the field's current state has been the prevalence of vision sensors in labs and consumer products, such as smartphones and vehicles. This not only allows large volumes of data to be collected for development and thorough testing but also means that solutions can be developed for deployment in various environments, e.g., in-cabin monitoring. Meanwhile, the introduction of convolutional neural networks (CNNs) \cite{lecun1989backpropagation} has allowed computers to better leverage the spatial properties of visual data and computationally scale with the size of the input image.

In common digital cameras, a grid of sensors is sampled according to some temporal sampling rate, and a frame is output consisting of the different values detected by each sensor in the grid. In contrast, event cameras (ECs), also known as dynamic vision sensors (DVSs), are novel bio-inspired sensors that record changes in scene brightness by generating "events" triggered by light intensity changes. Unlike traditional cameras, ECs produce a stream of asynchronous events in a text file rather than sampling the entire grid at a fixed rate \cite{shariff2024event}. Each event captures the polarity of the change along with spatial and temporal information. 

EC solutions require fast, real-time processing and analysis of event streams to avoid overwhelming the system with the sheer volume of captured events. The result is an energy-efficient visual sensor capable of ultra-low latency acquisition of visual information, with microsecond-level response capabilities \cite{shariff2024event}.

Research into potential applications of event data has touched upon many problem domains such as object detection \cite{gehrig2022pushing}, facial expression recognition \cite{barchid2023spiking}, and gaze estimation \cite{li2024gaze}, among others \cite{shariff2024event}. However, biomedical applications of ECs have yet to be significantly explored \cite{gallego2020event, shariff2024event, chakravarthi2024recent}, offering a valuable opportunity for research. The high dynamic range, low latency, and sensitivity to light present ECs as extremely suitable sensing technology for real-time health monitoring applications, such as remote heart rate estimation.

Cardiac pulse and any irregularities present in it hold valuable insight into an individual's health and well-being. However, as it is not always feasible or convenient for subjects to have on wearable heart sensors, research into contactless, non-invasive heart signal and heart rate monitoring has seen great interest over the past decade and a half \cite{verkruysse2008remote, liu2020multi, li2023non}. Contact-free heart signal monitoring is typically carried out using either remote-photoplethysmography (r-PPG), ballistocardiography (BCG), or both. As the blood's oxygen level changes, so do its light-reflecting properties, leading to subtle changes in skin color as blood is circulated. The process of remotely monitoring those subtle changes is known as r-PPG. BCG, on the other hand, is focused on subtle head motion resulting from the mechanical force of blood being pumped into the head \cite{liu2020multi}.

This study details the use of end-to-end convolutional neural networks to extract cardiac pulse from an EC's recording of the subject's face. Pre-existing frame-based methods \cite{liu2020multi} are leveraged by generating a 2-dimensional (2D) representation of temporally-binned events, referred to as an event frame. Additionally, the effect of decreasing the temporal bin size (i.e., increasing the frame rate) on the estimated heart rate is evaluated. As there exists no publicly available dataset that meets the study's requirements, the dataset used for this study's experiments was collected locally.

Our contributions are as follows:

\begin{itemize}
    \item To the best of the authors' knowledge, this is the first study on contact-free estimation of heart signals from event face data. As well as the first study to utilize CNNs for event-based vital metric monitoring.
    \item The study also proposes a novel method for pre-processing the cardiac signal sensor data used to annotate the event frames.
    \item An analysis of the effect decreasing the temporal bin period used to generate event frames has on the estimated heart rate.
\end{itemize}

The rest of the paper is structured as follows: Section \ref{sec:bckgrd} provides a brief overview of literature related to the task of remote cardiac pulse estimation and event cameras. Section \ref{sec:method} explains the methodology of the presented study from data preparation to model input and output processing. The experimental details and results are described in section \ref{sec:exp}. Section \ref{sec:disc} discusses the contributions presented, including a comparison with other methods from the literature, and highlights certain limitations of the presented study. Finally, section \ref{sec:conc} summarizes what was presented in this study and provides closing remarks on future work.

\section{Related Work} \label{sec:bckgrd}

\subsection{Remote Heart Rate Monitoring}
Early work on r-PPG has established that blood flow can be remotely measured on the human face using simple digital consumer cameras \cite{verkruysse2008remote}. This was done by extracting the heart signal from the video's green channel after spatially averaging the RGB video. The authors in \cite{de2013robust} investigated motion robustness when r-PPG is extracted from a colour video camera by developing and testing a number of potential algorithms that separate the blood pulse signal from motion-induced distortions using chrominance signals.

In contrast, deep learning methods are able to learn optimal weights for feature extraction, thus obviating the necessity for hand-crafted features and extensive pre-processing. This is demonstrated by DeepPhys in \cite{chen2018deepphys}, where a convolutional attention network (CAN) is presented for end-to-end extraction of cardiac pulse. The model's architecture consists of two branches, the motion branch and the appearance branch, each made up of several convolutional and pooling layers. The role of the motion branch is to extract complementary cardiac information based on minute body motion resulting from the mechanical flow of blood. Meanwhile, the appearance branch learns spatial attention masks \cite{vaswani2017attention} which are sent to the motion branch to compensate for the effect of illumination and other external factors. \cite{liu2020multi} built upon the work presented in \cite{chen2018deepphys} by introducing several CAN implementations, one of which being the temporal shift-CAN (TS-CAN). In TS-CAN, a temporal-shift module (TSM) is introduced before each convolutional layer in the motion branch. The TS function allows the exchange of temporal information between frames \cite{lin2019tsm}, thus enabling 2D convolutions access to temporal information.

In \cite{yu2019remote}, the authors propose two different implementations for another heart rate estimation end-to-end model, PhysNet. The first implementation utilizes 3-dimensional convolutions while the other uses a recurrent neural network (RNN) \cite{mcculloch1943logical} as the model backbone. The results presented showed that the 3-dimensional convolution implementation performed much better.

Another novel approach is presented in \cite{zou2024rhythmformer}, where the authors outline an end-to-end transformer-based method for extracting r-PPG signals, RhythmFormer. This approach manages to achieve state-of-the-art performance with fewer parameters, despite previous transformer-based approaches failing to outperform CNN methods significantly \cite{revanur2022instantaneous, yu2022physformer, yu2023physformer++}.

\subsection{Neuromorphic Event Camera for Pulse Estimation}
The applications of event streams and various representations of event data have been investigated for a wide variety of computer vision tasks \cite{gallego2020event, chen2020eddd, ryan2021real, ren2023spikepoint, shariff2024event}. However, to the best of the authors' knowledge, there exist only two studies that closely align with the proposed research question.

The authors in \cite{jagtap2023heart} explore the use of ECs to monitor the subject's wrist for pulse motion. A dark circle, approximately 1.63 cm in diameter, is drawn on the inside of the subject's wrist, near where the radial artery pulsates. This is done to create a high-contrast region which allows pulses to show up as a rapid series of events in the EC output. To calculate heart rate, event streams are transformed into a 1280 × 720 frame, and a heat map is used to find the 100 × 100 region of interest (RoI) with the highest sum of pixel activations. The RoI is then divided into smaller, non-overlapping 5 × 5 tiles for which the dominant frequency is obtained using a periodogram \cite{smeaton2023periodicity}; that frequency is the estimated pulse rate for the recording. 

While this study demonstrates the possibility of using ECs to capture arterial pulsation under a specific set of conditions, it has several critical limitations. A steady hand position, as mandated during their acquisition protocol, is difficult to maintain in most day-to-day events. Moreover, as their approach relies on marking the subject to create a RoI with high-contrast, the pulse monitoring is not completely contactless. Additionally, it is designed to detect visible wrist motion, in contrast to the more subtle physiological information present in the face region.

Meanwhile, the work presented in \cite{inproceedings} focused on monitoring the abdominal and chest area using ECs. Data was collected while the subjects were at rest and after they had conducted physical exercise on a treadmill. To reduce noise due to motion, subjects were instructed to rest their arms on their lap and lean against the chair's backrest. The captured events were binned into 10ms bins, and the number of events per bin was obtained. The signal composed of events-per-bin counts was passed through a Bandpass filter, and Fast Fourier Transform (FFT) \cite{bracewell1986fourier} was applied to the output to obtain the heart rate.

Their work further advances the field, demonstrating marker-free cardiac monitoring on an area of the body with more subtle physiological features. However, despite the positive reported results, this solution suffers from several issues. To start with, relying solely on event count without any spatial information restricts the use case as scene information is removed. This is further demonstrated by the strict position required from the subjects. Moreover, this study was only tested on a total of 10 minutes' worth of data from 10 subjects (two 30-second sessions each). To validate this method's viability for unconstrained heart rate estimation from events, we implement the algorithm per the authors' description and test it on the face-based dataset used for the presented study. The results presented in section \ref{sec:disc} demonstrate its unsuitability for this task.

The presented study investigates pulse signal detection from event streams of the subjects' faces. No additional sensors or markings are required, and the feature extraction by the end-to-end model allows for larger freedom of movement for the subjects. Moreover, relying on spatial features as well as both motion and light changes allows the proposed method to be much more robust than approaches based solely on event counts.

\section{Methodology} \label{sec:method}
\begin{figure*}[tb]
  \centering
  \includegraphics[height=7cm]{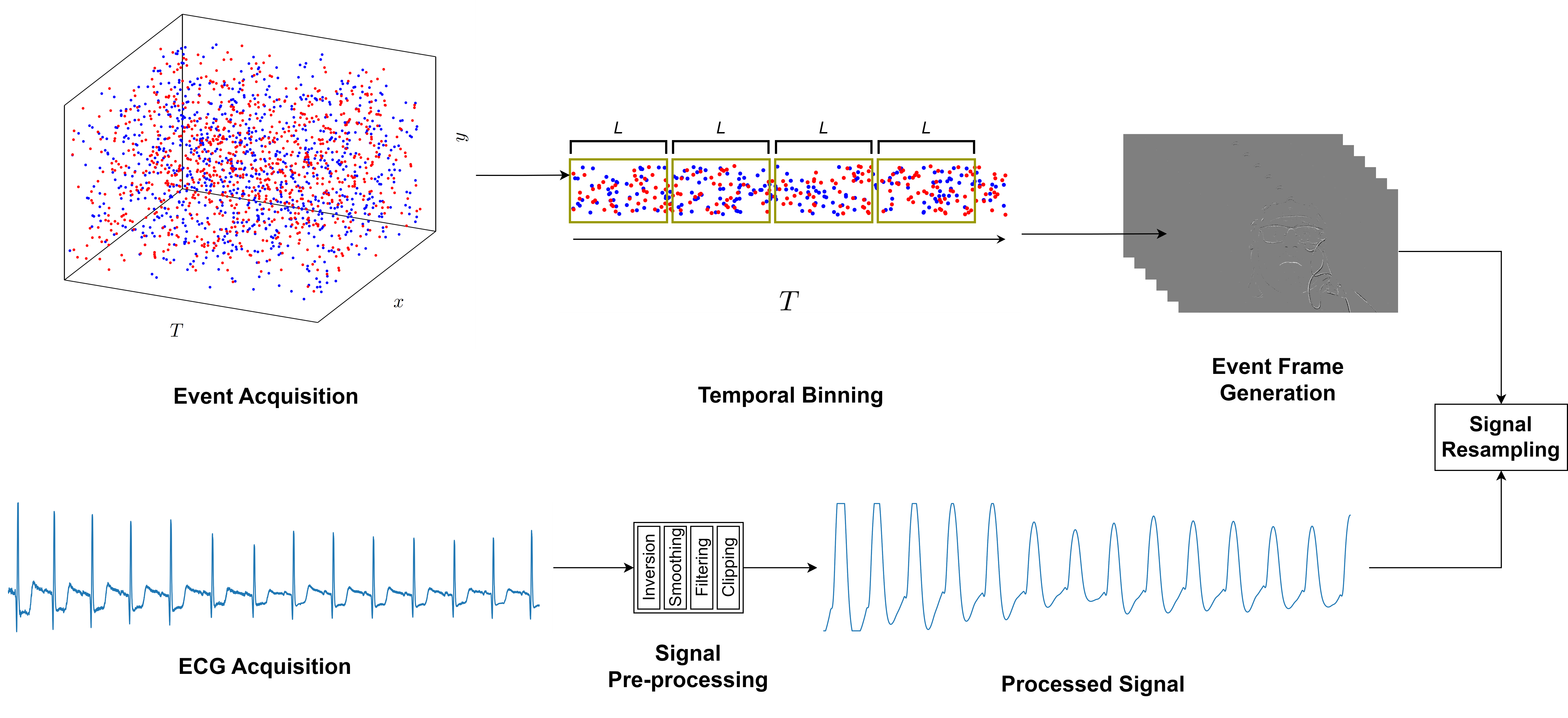}
  \caption{Preparation of time event data for remote cardiac pulse estimation. The acquired events are divided into fixed-sized windows of period L, and an event frame is generated for each window. Meanwhile, the corresponding ECG signal is processed using a multi-step algorithm and then resampled based on the selected frame rate for the generated event frames. This produces the frame-label pairs used for the supervised training of the neural network.}
  \label{fig:method}
\end{figure*}

The methodology proposed in this study is outlined in Fig. \ref{fig:method}. Visual and health sensor data are acquired simultaneously and then used for supervised training of the end-to-end convolutional neural model. The event stream is divided into non-overlapping temporal windows/bins, each covering a fixed period $L$, which are each used to generate the 2D event frame inputs. The cardiac pulse sensor signal is passed through a signal processing algorithm composed of inversion, smoothing, filtering, and clipping. Event and health signal timestamps are then used to resample the processed signal based on the chosen event window width $L$, and a standard CNN model is trained using the frames and corresponding cardiac signal labels.

\subsection{Input Pre-processing}
\subsubsection{Event Pre-processing}
The event stream is composed of a series of 4-dimensional events $e_i = (t_i, x_i, y_i, p_i)$ where $e_i$ is the $i$-$th$ event and $t_i, x_i, y_i,$ and $p_i$ are the timestamp (typically in microseconds), x-coordinate, y-coordinate, and polarity of $e_i$. For frame generation, those events are broken into non-overlapping windows such that, for the $j$-$th$ event window $W_j$:

\begin{equation}
W_j \{e_i, e_k\} \mid j\in[1, n_{win}], t_k - t_i = L, i < k
\end{equation}

where $L$ is a fixed time period, $n_{win}$ is the total number of windows, and $i$ and $k$ are the indices of the first and last events in $W_j$.

The events used for the experiments in section \ref{sec:exp} were captured by a Prophesee event camera \cite{lichtsteiner2008128}, which generates events with polarities of 0 or 1. Before frame generation, all 0-polarity events have their polarities mapped to -1 to distinguish negative events from lack of events. 

To obtain a square face-crop, fixed-index cropping is applied directly to the events in each bin/window. This involves eliminating events taking place outside a fixed set of $x$ and $y$ boundaries and shifting the $x$-coordinates of the remaining events to the left. For computational efficiency, the events in each window are then spatially downsampled by dividing the corresponding $x$ and $y$ coordinates by downsampling factor $d_f$. 

Following that, a 2D event frame is generated for each window $W_j$ through the accumulation of event polarities for events triggered in the same $x$-$y$ coordinates. This stage is carried out such that for $idx \in [0, x_{max}y_{max}/d_f^2)$, the pixel value $P_{idx}$ is as follows:

\begin{equation}
P_{idx} = \sum_{n \in (i, k)}{}{e_n} \mid (x_n + y_n * x_{max} / d_f) = idx
\end{equation}

where $x_{max}$ and $y_{max}$ are the highest possible values for the raw events' $x$ and $y$ values, and $idx$ is used to index the frame pixels being generated using a single coordinate. The resulting array of pixels is then reshaped into a 2D frame.

The $n_{win}$ generated frames have their pixel values normalized between -8 and 8, which was found to be best-practice for visual clarity and noise reduction. Finally, the unsigned integer event frames used as model input are obtained by multiplying those frames by 255, followed by typecasting into 8-bit unsigned integers.

\subsubsection{Standard Camera Frame Pre-processing}
RGB remote pulse estimation is treated as the baseline for this task; therefore, a comparison between models trained on event and RGB frames is vital. 

A fixed-index face cropping is applied on the RGB frames; this crop is then spatially downsampled using pixel area relation resampling. Downsampling helps reduce camera quantization error and improve computational complexity at the cost of losing spatial resolution. The DeepPhys architecture authors \cite{chen2018deepphys} presented a discrete approximation of low-noise light-invariant motion representation $N(t)$ using normalized frame difference of the downsampled frames $D(t)$ by subtracting two consecutive frames and dividing the difference by their sum. This representation is used as input to the motion branch of the TS-CAN model used for the experiments in section \ref{sec:exp}, with z-standardized frames used as input to the appearance branch.

\subsection{Label Pre-processing}
\begin{figure}[tb]
  \centering
  \includegraphics[height=8cm]{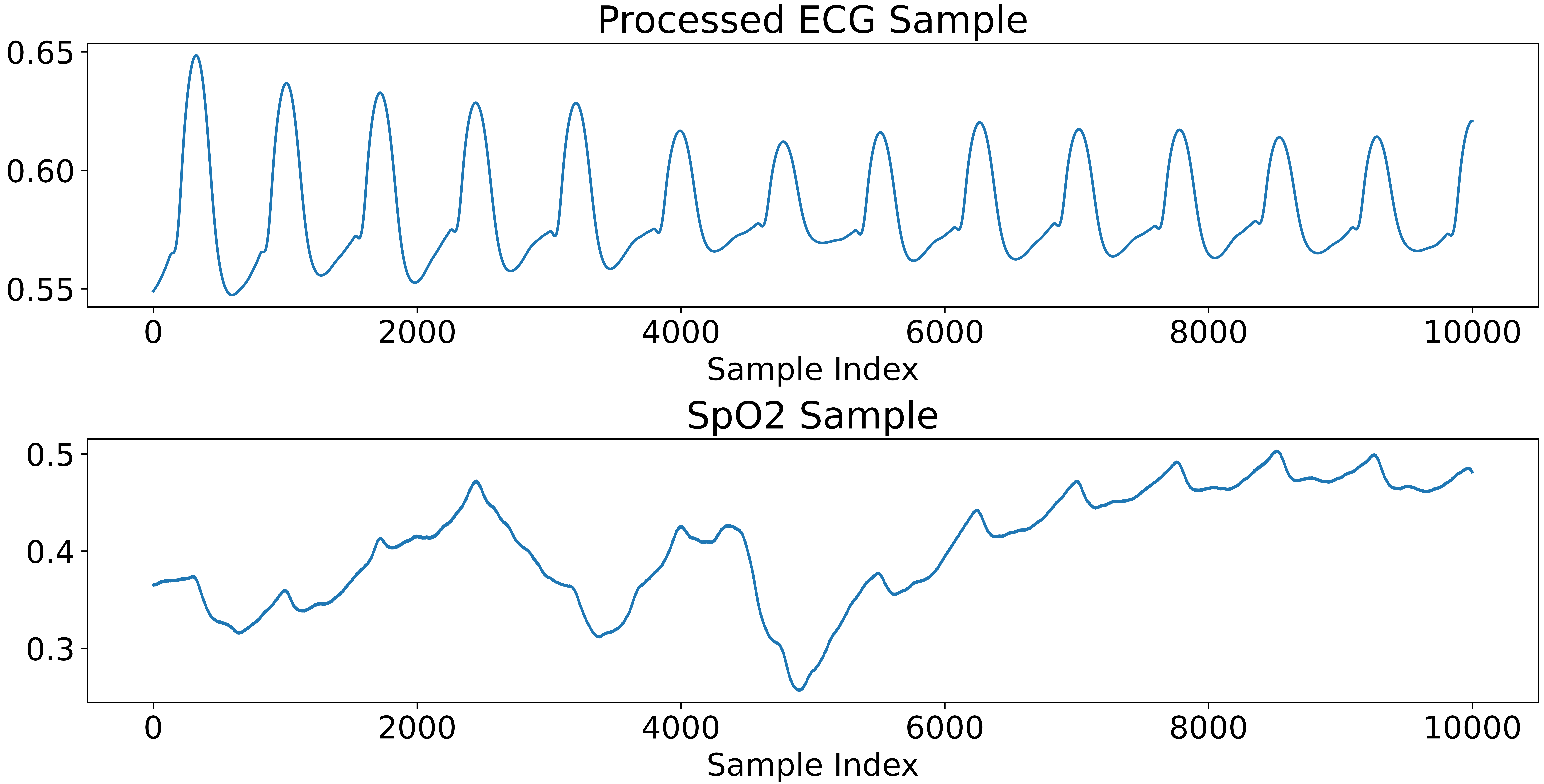}
  \caption{Comparison between ECG and SpO2 data acquired during the same period at a sampling rate of 1 kHz. The SpO2 signal exhibits high motion noise during the first half of the acquisition, while the ECG signal remains unchanged.}
  \label{fig:spo2}
\end{figure}

An electrocardiogram (ECG) sensor was chosen as the source for the cardiac signal labels corresponding to each frame. ECG sensors are attached to the torso and are more robust than SpO2 sensors against noise and motion artifacts. A 10-second sample is graphed in Fig. \ref{fig:spo2} to demonstrate an example of the kind of noise SpO2 data is prone to. 

Typically, when ECG data is used to generate image labels for supervised training, the signal is processed to resemble SpO2 signals more closely \cite{jaiswal2022heart, yu2023physformer++}. This is because both the SpO2 signal and the remote signal extracted from the face measure the change in blood in the extremities, while the ECG sensor monitors electrical heart activity \cite{yu2019remote}.

\begin{figure*}[tb]
  \centering
  \includegraphics[height=15cm]{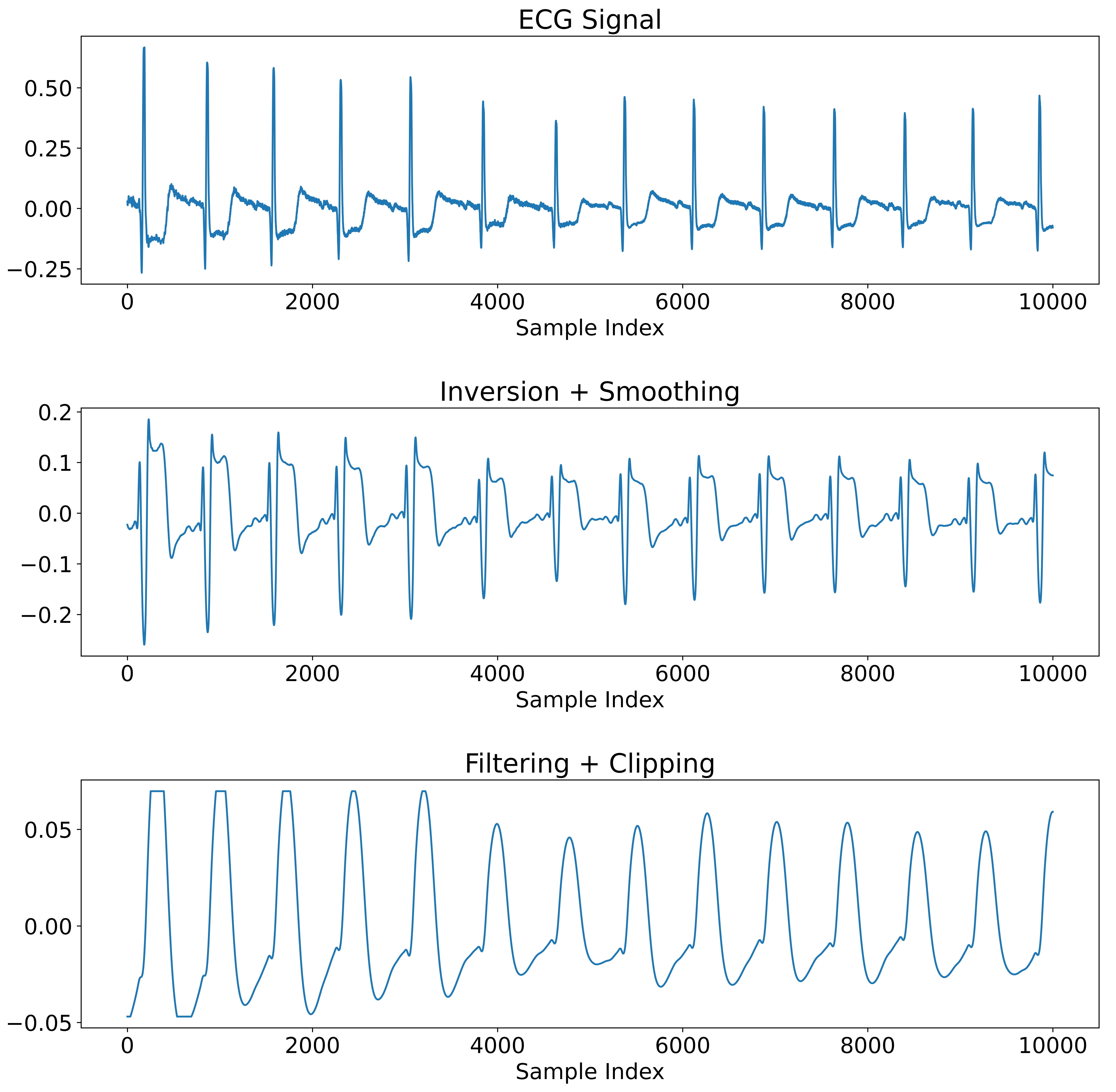}
  \caption{Illustration of pre-processing steps applied to the ECG signal to obtain the signal used to label frames. A 10-second sample is inverted and smoothed. This is followed by Bandpass filtering and clipping of extreme values.}
  \label{fig:ecg_proc}
\end{figure*}

A multi-step algorithm is proposed to transform the raw ECG signal into a form more suitable for supervised training labeling. This approach incorporates commonly used steps such as bandpass filtering \cite{jaiswal2022heart} and smoothing \cite{yu2023physformer++}, while combining them with additional steps to better align the signal with changes within the RoI. The step-by-step algorithm outputs are shown using a 1-second ECG sample in Fig. \ref{fig:ecg_proc}.

\begin{figure}[tb]
  \centering
  \includegraphics[height=5cm]{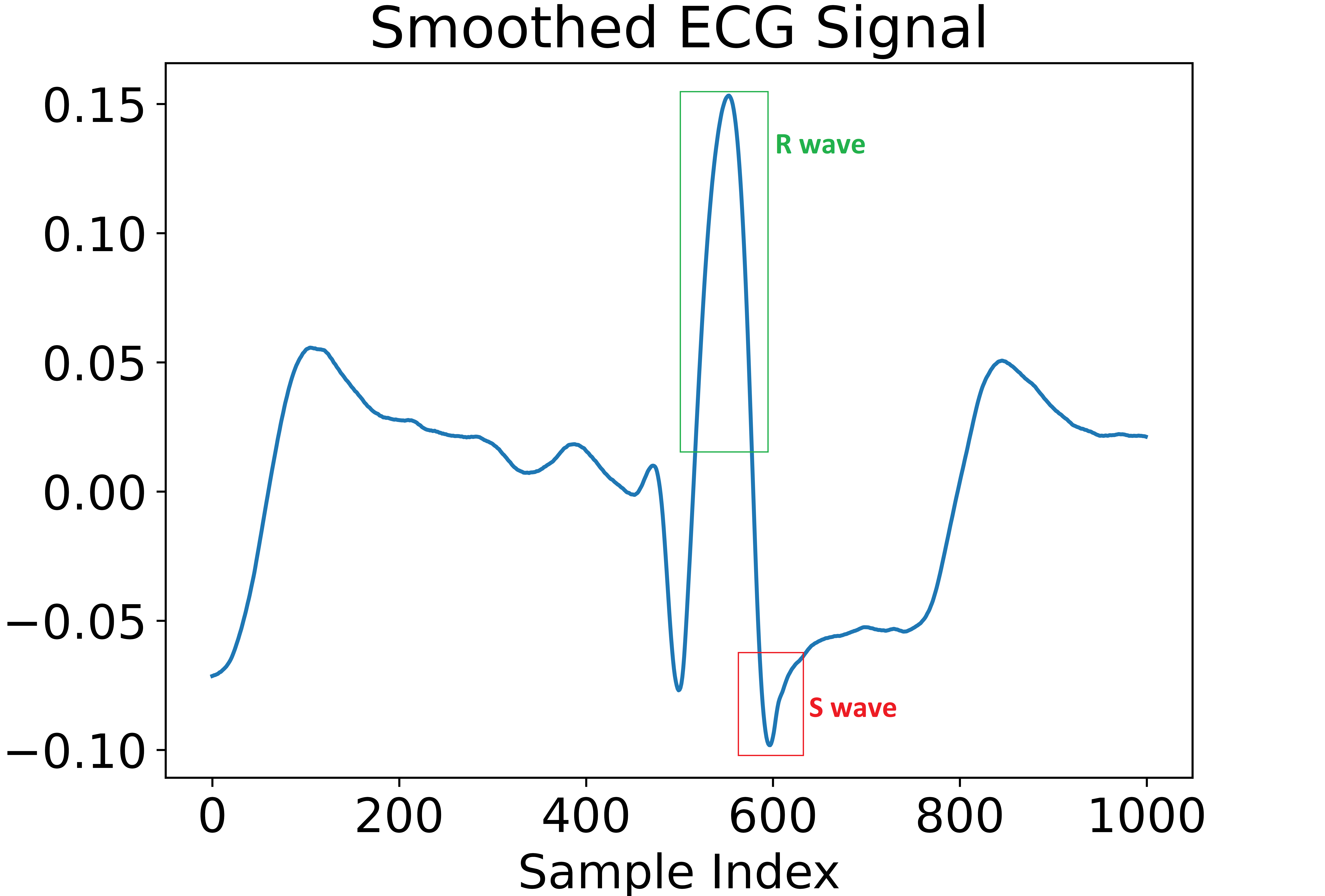}
  \caption{The R and S waves on a standard ECG peak.}
  \label{fig:ecg_rs}
\end{figure}

The ECG sensor data corresponding to each event stream is inverted (so that the r-wave is pointing downwards) and then further processed as outlined in Fig.\ref{fig:method}. The inverted signal is used to focus on detecting s-wave peaks (Fig. \ref{fig:ecg_rs}) to compensate for the delay present between each heartbeat (r-wave) and the time it takes oxygenated blood to reach the extremities (e.g., the face).

For the remaining signal pre-processing steps, several parameters are adjusted according to the sensor's sampling rate. For an ECG signal sampled at 1 kHz, a Savitsky-Golay filter \cite{savitzky1964smoothing}, with a window size of 101 and a quadratic fitting equation, is used for smoothing. A first-order Bandpass Butterworth filter with 0.0015 and 0.0048 critical frequencies (based on a sampling rate of 1 kHz and standard heart rate cutoff frequencies of 0.75 Hz and 2.5 Hz) is applied to the smoothed signal. Following that, the signal values in the top and bottom 1\% of the signal are clipped to eliminate outliers.

The signal is then resampled based on the frame rate of the event or RGB frames. Timestamps are used to map a signal value to the temporally closest frame. For event frames, the timestamp of the last event in each event window is used as the timestamp of that frame. The resulting resampled signal is then prepared for supervised model training by calculating the first derivative as recommended in \cite{chen2018deepphys}, followed by the mean and standard deviation being normalized to zero and one, respectively.

\subsection{Output Post-processing}
Before being used to calculate heart rate, the predicted signal is first passed through a standard r-PPG post-processing pipeline \cite{liu2022deep} to remove noise and emphasize the relevant features.

During the first step, a cumulative sum function is applied to the output signal, as the model is trained to estimate the normalized signal difference. Following that, the signal is detrended to reduce noise using the implementation provided in the rPPG-toolbox \cite{liu2022deep}. A first-order Bandpass Butterworth filter is then applied to the detrended signal, and the output is subsequently used to calculate the heart rate for each subject via FFT \cite{bracewell1986fourier}.

The differences between the ground-truth and predicted heart rates, denoted as ${d}_{hr}$, are used to compute the model's error value.

\section{Experiments} \label{sec:exp}
\subsection{Data}
The data used for this study is one phase from a large-scale acquisition focused on physiological and psychological remote health monitoring. Informed consent was obtained from all the subjects, and the acquisition was approved by an internal ethics committee.

Before the start of the acquisition, a number of health sensors were attached to the subjects. The sensors included an electroencephalogram (EEG) head cap, an electrooculography (EOG) sensor around the eyes, a pulse oximetry (SpO2) sensor on the fingertips, and an ECG sensor on the chest area. While the EEG and EOG sensors might obscure certain regions of the face, the data they capture is valuable for health monitoring; however, that is beyond the scope of this study.

After placing the sensors, subjects were instructed to sit in front of an array of cameras, including a 1280 × 720 Prophesee \cite{lichtsteiner2008128} EC and an 800 × 600 Brio standard RGB camera capturing at a rate of 30 frames per second (FPS). During this acquisition phase, subjects were told to face forward and keep their eyes closed for two minutes, then keep them open for another two minutes, followed by a one-minute break. Some subjects chose to remain within the acquisition parameters during their break, thus providing an additional minute of viable data. Heart signal data was collected using both a Biosignals Plux \cite{pluxbiosignals} ECG sensor and a SpO2 sensor at 1 kHz.

The dataset used in this study consisted of data from a total of 64 subjects (45 male, 19 female). Most subjects (46 subjects) had four minutes worth of data included in the dataset, while a minority had three (9 subjects) or five (9 subjects) minutes of data. Subjects with face occlusions, such as beards or glasses, were included in the dataset.

The window period and downsampling factor were set to $L = 33333$ $\mu s$ and $d_f = 5$, thus generating 144 × 144 30FPS event frames. For the RGB frames, fixed-index cropping is applied to get a 400 × 400 face crop, which was downsampled to 144 × 144.

\subsection{Neural Model}
Model selection required a choice that is capable of extracting physiological information from facial features while being robust to varying illumination and head motion. One class of end-to-end models that have shown reliable performance for this use case are CAN models, with TS-CAN \cite{liu2020multi} demonstrating the best real-time performance relative to similar architectures. The DeepPhys two-branch feature extractor was enhanced by introducing temporal-shift modules \cite{lin2019tsm} to the motion branch. This enables the model to leverage facial features, motion, and temporal information while being robust against varying illumination configurations. The frame depth parameter controls the number of frames the temporal shift function is applied on and was set to 10 for the presented experiments.

The base network and training implementation are obtained from the rppg-toolbox repository \cite{liu2022deep}, with any configurations not mentioned in this section being kept at their default values. For the RGB model, as explained in section \ref{sec:method}, motion representation $N(t)$ is used as input for the first model branch. For the second branch, z-score standardization is carried out on the downsampled frames $D(t)$. For faster convergence, both inputs were scaled to unit standard deviation and a mean of zero.

\begin{figure}[tb]
  \centering
  \includegraphics[height=8cm]{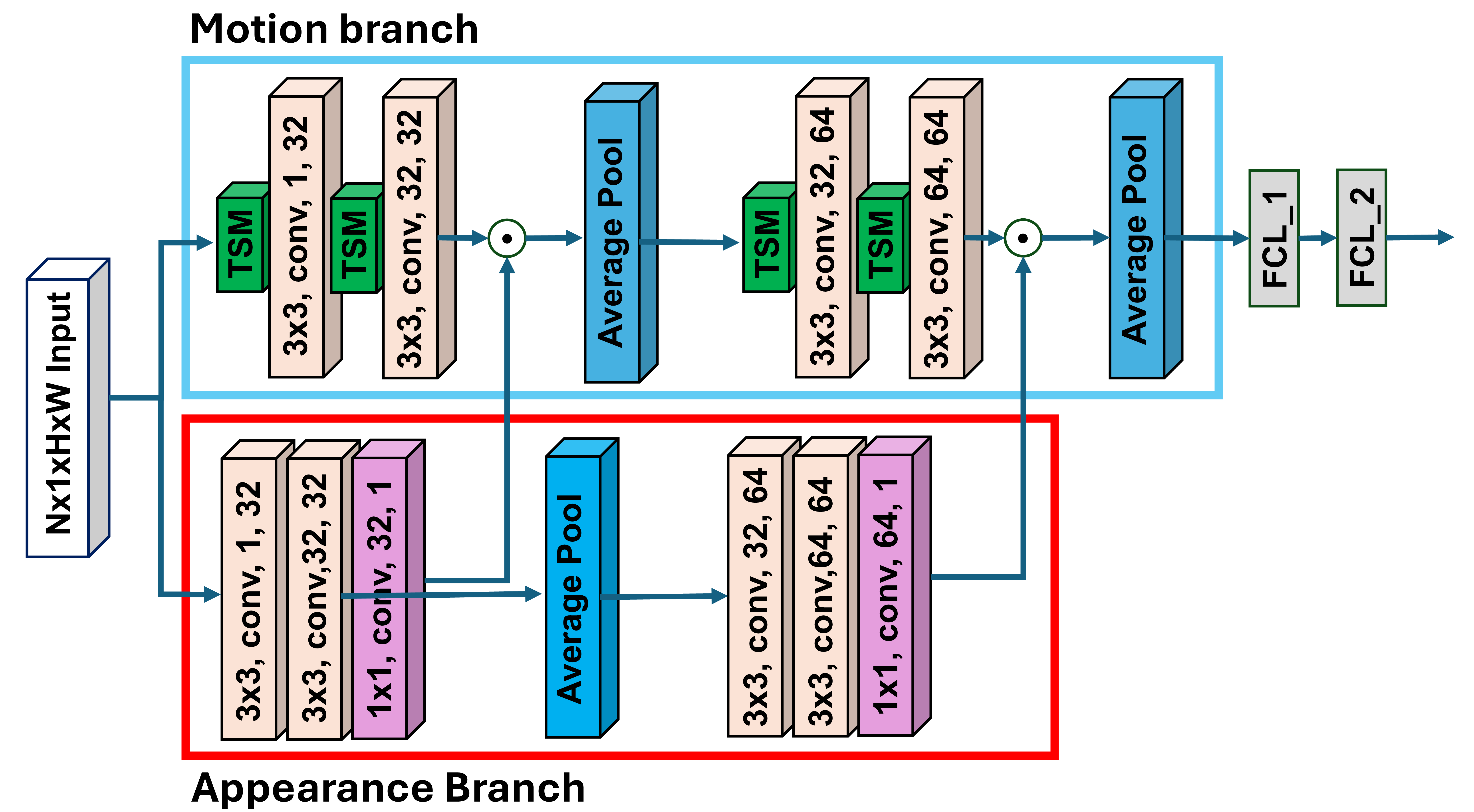}
  \caption{Diagram of the modified TS-CAN trained on event frames. As opposed to the original TS-CAN model where each branch takes in a different 3-channel input, the branches of this variant both take in the same 1-channel event frame input. The temporal shift modules (TSM) in the motion branch and the element-wise multiplication attention mechanism follow the original design.}
  \label{fig:event_tscan}
\end{figure}

A modified version, illustrated in Fig. \ref{fig:event_tscan}, is used for the event data experiments. Instead of frame differences and standardized frames as input to one branch each, each branch is modified to take in the single-channel event frames.

The training code used and the trained CNN model weights are available in a GitHub repository \footnote{\url{https://github.com/C3Imaging/Contactless_Cardiac_Pulse_Monitoring_Using_Event_Cameras}}.

\subsection{Training Configuration}
The training set consisted of 44 subjects, with 10 subjects randomly selected for the validation and testing sets. All models were trained for 30 epochs, using validation loss to select optimal weights to test on the test set.

An AdamW optimizer \cite{loshchilov2017fixing} is used along with the mean square error (MSE) loss function. A OneCycleLearningRate learning rate scheduler \cite{smith2019super} was employed with the initial learning rate as the maximum learning rate. As a learning rate scheduler was used, the epoch size also affected the learning rate. Both models converged before the maximum number of epochs. 

Data augmentation was introduced through random horizontal flips with a 50\% probability. Both models were trained using a learning rate of $18e-5$ and a batch size of $b=8$. With regards to batch size, each "item" corresponded to a chunk of 180 frames; thus, each batch of size $b$ contained $180*b$ frames and labels.

Training was carried out on an Ubuntu workstation utilizing an i9-13900KS CPU and two RTX A6000 Nvidia GPUs.

\subsection{Experimental Results} \label{sec:res}
Four performance metrics were selected to evaluate the model-predicted signal based on beats per minute (bpm) heart rate error: mean absolute error (MAE), root mean squared error (RMSE), mean absolute precision error (MAPE), and Pearson correlation coefficient. The heart rate estimation results for both models are summarised in Table \ref{tab:tbl1}. 

\begin{table}[ht]
    \caption{Experimental Results of Heart Rate Estimation on Local Data} \label{tab:tbl1}
    \centering
    \setlength{\tabcolsep}{5pt}
    {\renewcommand{\arraystretch}{1.5}
    \begin{tabular}{c c c c c}
        \hline
        \hline
        Frame Type & MAE & RMSE & MAPE & Pearson\\
        \hline
        \hline
        RGB & \textbf{1.89} & \textbf{2.92} & \textbf{2.51} & \textbf{0.95}\\
    Events & 2.33  & 3.32 & 3.02 & 0.93\\
    \hline
    \end{tabular}}
\end{table}

Overall, the model trained on RGB frames performed better, with a gap of 0.4 - 0.51 bpm across the mean error metrics. Despite that, the heart rate extracted from the event data was within the acceptable error range for r-PPG methods ($\leq$ 5 RMSE) \cite{poh2010non}. Further testing details, in the form of per-subject heart rate RMSE, are presented in Table \ref{tab:tbl2}. The test set consisted of 9 male and 1 female (subject 5) subjects, with 3 subjects wearing glasses (subjects 0, 1, and 9), and 1 with a beard (subject 1).

\begin{table}[ht]
    \caption{Heart Rate Estimation Root Mean Squared Error for Each Subject} \label{tab:tbl2}
    \centering
    \setlength{\tabcolsep}{5pt}
    {\renewcommand{\arraystretch}{1.5}
    \begin{tabular}{c c c c c}
        \hline
        \hline
        Subject ID & Occlusions & RGB & Events\\
        \hline
        \hline
        0 & Glasses & \textbf{0.00} & 0.44\\
        1 & Glasses, Beard & \textbf{0.00} & 7.69\\
        2 & None & 1.98 & \textbf{0.00}\\
        3 & None & 5.27 & \textbf{0.00}\\
        4 & None & \textbf{1.54} & 2.86\\
        5 & None & \textbf{0.00} & 2.42\\
        6 & None & \textbf{0.00} & 3.08\\
        7 & None & \textbf{5.05} & \textbf{5.05}\\
        8 & None & 5.05 & \textbf{1.32}\\
        9 & Glasses & \textbf{0.00} & 0.44\\
    \hline
    \end{tabular}}
\end{table}

Both models achieved the same results on 1 subject. Meanwhile, the RGB model outperformed the events model on 6 subjects and performed worse on 3 subjects.

To further explore the impact of frame rate on the events model performance, two more models were trained on frames generated using periods of $L = 16666$ $\mu s$ and $L = 8333$ $\mu s$ (i.e., 60 FPS and 120 FPS). The processed ECG signal (originally sampled at 1 kHz) was downsampled accordingly.

With regards to model training, the frame depth parameter was set to 20 frames for the model being trained on the 60 FPS data, and 40 frames for the model trained on the 120 FPS data. For both experiments, the learning rate was halved to $9e-5$. The overall heart rate performance of the four models is presented in Table \ref{tab:tbl3}.

\begin{table}[ht]
    \caption{Heart rate estimation performance for models trained on RGB frames and event frames generated using different periods. The best results are in bold, with the second-best results underlined.} \label{tab:tbl3}
    \centering
    \setlength{\tabcolsep}{5pt}
    {\renewcommand{\arraystretch}{1.5}
    \begin{tabular}{c c c c c c}
        \hline
        \hline
        Frame Type & FPS & MAE & RMSE & MAPE & Pearson\\
        \hline
        \hline
        RGB & 30 & \underline{1.89} & 2.92 & \underline{2.51} & \underline{0.95}\\
    Events & 30 & 2.33  & 3.32 & 3.02 & 0.93\\
    Event & 60 & 2.18 & \underline{2.54} & 2.76 & \underline{0.95}\\
    Event & 120 & \textbf{1.58} & \textbf{2.13} & \textbf{2.20} & \textbf{0.97}\\
    \hline
    \end{tabular}}
\end{table}

The increased frame rate has a noticeable positive effect on the event frame models' heart rate estimation performance. The model trained on the 60 FPS event frames manages to achieve performance similar to that of the RGB model. Meanwhile, the 120 FPS event frames model outperforms the other three models across all metrics. The per-subject RMSE results are presented in Table \ref{tab:tbl4}.

\begin{table}[ht]
    \caption{Heart rate estimation root mean squared error of the four models for each subject. Results exceeding the acceptable error rate are highlighted in red, with the best results per subject highlighted in bold.} \label{tab:tbl4}
    \centering
    \setlength{\tabcolsep}{5pt}
    {\renewcommand{\arraystretch}{1.5}
    \begin{tabular}{c c c c c c}
        \hline
        \hline
        Subject ID & RGB & Events 30 FPS & Events 60 FPS & Events 120 FPS\\
        \hline
        \hline
        0  & \textbf{0.00} & 0.44 & 1.76 & 0.66\\
        1  & \textbf{0.00} & \textcolor{red}{7.69} & 1.32 & 1.54\\
        2  & 1.98 & \textbf{0.00} & 3.96 & \textbf{0.00}\\
        3 & \textcolor{red}{5.27} & \textbf{0.00} & 0.22 & 2.42\\
        4 & \textbf{1.54} & 2.86 & 3.30 & 2.42\\
        5 & \textbf{0.00} & 2.42 & 1.76 & \textcolor{red}{5.05}\\
        6 & \textbf{0.00} & 3.08 & 1.10 & 1.76\\
        7 & \textcolor{red}{5.05} & \textcolor{red}{5.05} & 4.61 & \textbf{0.00}\\
        8 & \textcolor{red}{5.05} & \textbf{1.32} & \textbf{1.32} & \textbf{1.32}\\
        9 & \textbf{0.00} & 0.44 & 2.42 & 0.66\\
    \hline
    \end{tabular}}
\end{table}

\begin{figure*}[tb]
  \centering
  \includegraphics[height=15cm]{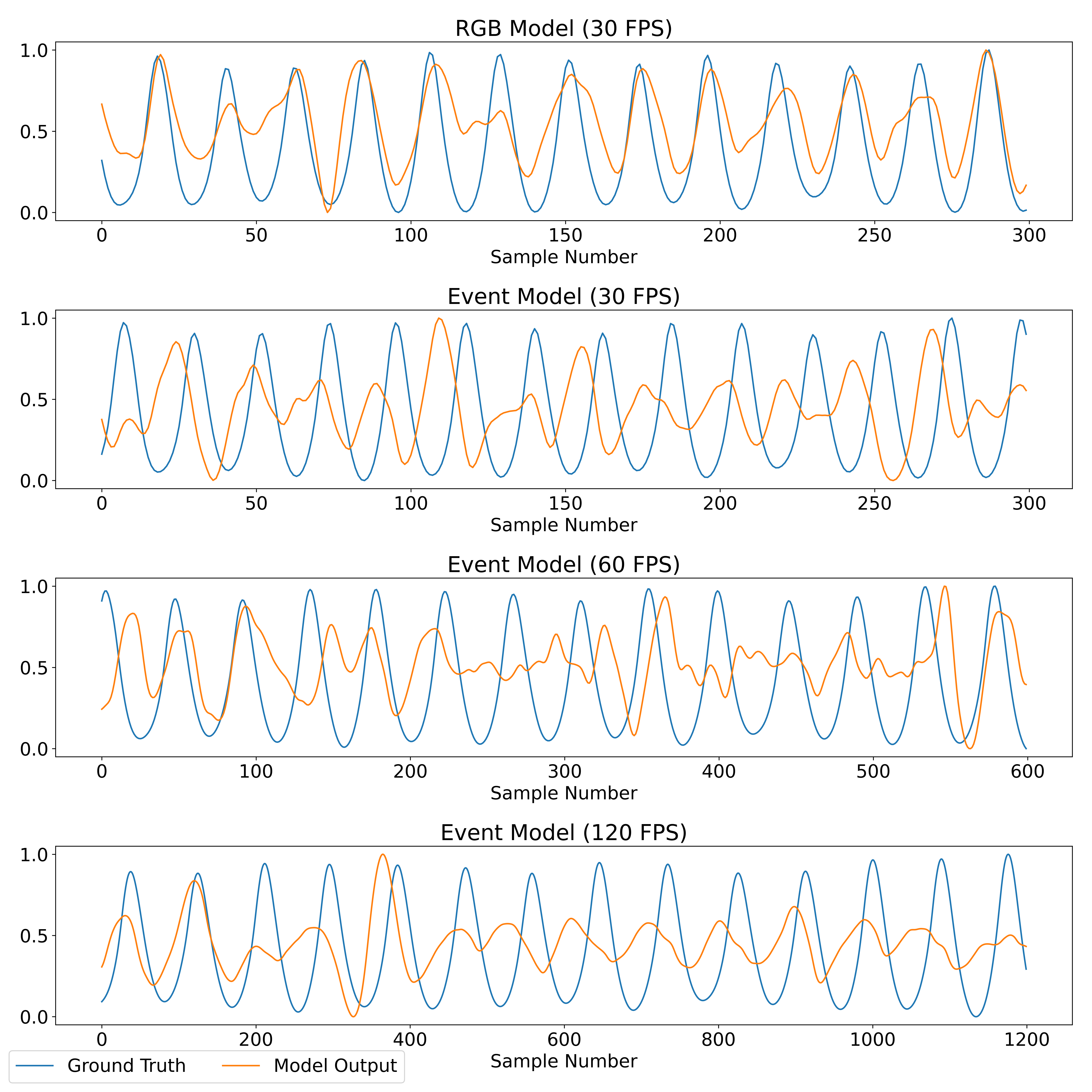}
  \caption{Sample output from the different r-PPG models covering a ten-second duration. Both the model output (orange) and ground-truth (blue) are post-processed and then normalized for clearer visual comparison.}
  \label{fig:model_comp}
\end{figure*}

Despite the overall higher error metrics, the RGB model performs better on a larger number of subjects than the event frame models, with all but one of those results involving a perfect estimation of heart rate. Meanwhile, the 30 and 120 FPS event models achieve the best results on three subjects each, with the 60 FPS event model only outperforming the other models on a singular subject.

In contrast, the RGB model similarly has the highest number of subjects with an error rate exceeding the 5 bpm RMSE threshold, followed by the 30 FPS events model and the 120 FPS events model. The only model that managed to successfully estimate the heart rate for each subject within an acceptable error rate was the 60 FPS event frame model.

A sample output covering ten seconds from each model for subject 7 is presented in Fig. \ref{fig:model_comp} and compared to the ground truth. A key observation is that, despite having the worst heart rate performance, the RGB model sample output aligns quite closely with the ground truth, in contrast to the 60 FPS events model which has a slightly better heart rate RMSE. In fact, the RGB and 120 FPS event frame models are the only ones that do not miss a single beat during the presented 10-second period, despite their heart rate RMSE being significantly different. This is because point-by-point (i.e., signal-based) loss does not always paint the full picture of a model's heart rate estimation performance \cite{mohamed_moustafa_2023_8309793}.

While the ECG sensor measures electrical heart activity rather than the actual motion of blood, the difference in alignment between the RGB and event frame graphs is likely due to the difference in the underlying data captured by each representation. Standard frames capture the state of the subject at each particular instant. In contrast, event frames include the entire change in the RoI state during each temporal window, encoding information from the entire time window rather than just information at a particular instant.

\section{Discussion and Limitations} \label{sec:disc}
The results reported in section \ref{sec:exp} outline the capabilities of ECs to estimate a subject's heart rate remotely through face-based recordings. Specifically, the results prove that ECs can successfully and reliably capture the underlying physiological information present in the face region. Another important finding is that near-RGB performance can be reached without relying on complex data representation or custom network architectures. Pre-existing research based on standard camera frames can be leveraged to develop biomedical EC solutions through simple event-frame representation.

\begin{table}[ht]
    \caption{Comparison of presented results with heart rate extracted through event count algorithm proposed in \cite{inproceedings}.} \label{tab:tbl5}
    \centering
    \setlength{\tabcolsep}{5pt}
    {\renewcommand{\arraystretch}{1.5}
    \begin{tabular}{c c c c c c}
        \hline
        \hline
        Bins/second & Event & MAE & RMSE & MAPE & Pearson\\
        \hline
        \hline
        \multirow{2}{*}{30} & Counts\cite{inproceedings} & 16.63 & 25.35 & 18.33 & -0.35\\
        & Frames(Ours) & \textbf{2.33} & \textbf{3.32} & \textbf{3.02} & \textbf{0.93}\\
        \hline
        \multirow{2}{*}{60} & Counts\cite{inproceedings} & 19.04 & 28.52 & 21.07 & -0.16\\
        & Frames(Ours) & \textbf{2.18} & \textbf{2.54} & \textbf{2.76} & \textbf{0.95}\\
        \hline
        \multirow{2}{*}{120} & Counts\cite{inproceedings} & 16.76 & 25.68 & 18.41 & -0.55\\
        & Frames(Ours) & \textbf{1.58} & \textbf{2.13} & \textbf{2.20} & \textbf{0.97}\\
    \hline
    \end{tabular}}
\end{table}

Table \ref{tab:tbl5} compares the results of the proposed methodology with the algorithm proposed in \cite{inproceedings} for event-based remote heart rate estimation. The raw events are binned using a sampling rate of 30, 60, and 120 bins per second to match the rates used in section \ref{sec:exp}. The event count for each bin is then obtained, and the event count signal is used to estimate the heart rate for each subject as described in \cite{inproceedings}. Overall, the performance of this algorithm fails to compete with that of the presented method. Not only is the error rate more than 8 times that achieved by the event frame-based method, but their method also fails to achieve an acceptable error rate across all sampling rates.

\begin{table}[ht]
    \caption{Comparison of presented results with heart rate extracted through wrist-based event recording. The best result for each metric is highlighted in bold, with the second-best underlined.}\label{tab:tbl6}
    \centering
    \setlength{\tabcolsep}{5pt}
   {\renewcommand{\arraystretch}{1.5}
    \begin{tabular}{c c c c c}
        \hline
        \hline
        RoI & Frame Type & Scenario & MAE & RMSE\\
        \hline
        \hline
        \multirow{2}{*}{Wrist \cite{smeaton2023periodicity}} & \multirow{2}{*}{Events} & After Rest & \textbf{1.56} & \underline{2.14}\\
        & & After Exercise & 1.63 & 2.21\\
         \hline
        \multirow{4}{*}{Face (Ours)} & RGB (30 FPS) & \multirow{4}{*}{Rest} & 1.89 & 2.92\\
        & Events (30 FPS) & & 2.33 & 3.32\\
        & Events (60 FPS) & & 2.18 & 2.54 \\
        & Events (120 FPS) & & \underline{1.58} & \textbf{2.13}\\
    \hline
    \end{tabular}}
\end{table}

Table \ref{tab:tbl6} compares the results of this study with the previous work done on remote pulse extraction from wrist monitoring \cite{smeaton2023periodicity}. Due to the inability to obtain an implementation of their presented method, their reported results are used for comparison instead. Overall, the presented results are similar to the results achieved through face monitoring at 120 FPS. The method proposed in that study requires the use of markings to enhance skin contrast in addition to monitoring a region of the body where the subject's pulse is much more physically visible (compared to the face). Yet, despite that, the methodology presented in this study can achieve similar performance while being fully non-invasive and much more suitable for real-life scenarios (e.g., driver monitoring) compared to the stringent requirements mentioned in their study.

The presented study, however, is still subject to certain limitations. First, the lack of relevant publicly available datasets necessitated the use of a proprietary dataset. Despite that, the study's main contributions are clearly outlined in ample detail in sections \ref{sec:method} and \ref{sec:exp}, and thus can be easily implemented using other event datasets. Secondly, the dataset used only includes stationary subjects. While this is a common scenario found in popular benchmark datasets such as the University Bourgogne Franche-Comté Remote PhotoPlethysmoGraphy (UBFC-rPPG) dataset \cite{bobbia2019unsupervised}, other r-PPG datasets such as the Pulse Rate Detection (PURE) dataset \cite{stricker2014non} typically include scenarios involving more interactions or movement. Since the goal of this study was to investigate the preservation of physiological information within event data as well as the feasibility of applying r-PPG methods on event frames of the face region, data captured using the scenario described in section \ref{sec:exp} was deemed most suitable for addressing the aforementioned research questions. Moreover, despite being asked to sit facing forward, subjects were not required to hold a specific pose, unlike in the data used in \cite{inproceedings}.

\section{Conclusion and Future Work} \label{sec:conc}
In conclusion, this paper investigates the use of time events for the remote detection of pulse signals. To our knowledge, this is the first work done on contact-free heart rate monitoring using face event data, as well as the first to use end-to-end models. A multi-step ECG processing algorithm is also presented for annotating model frame inputs. 

The event-frame model was able to achieve similar performance to the model trained on RGB frames, with the difference in RMSE being 0.4 bpm when both frames were sampled at a rate of 30 FPS. Models trained on event frames generated at rates of 60 FPS and 120 FPS managed to perform equally as well or better than the RGB model, respectively.

This work utilized event frames, a 2-dimensional representation of events. A survey of other representation methods and their performance on this task would be a valuable addition to the literature, as well as allow for compatibility with other architectures (e.g., graph networks or SNN). Event de-noising is another crucial process that needs to be integrated into the framework, as heart rate monitoring relies on detecting subtle changes in color and motion \cite{liu2020multi}. An investigation into the preservation/de-noising/enhancement of physiological information in synthetic events is warranted, as it would allow researchers to leverage the vast amount of public frame-based heart rate datasets \cite{xiao2024remote} for the development of EC applications.

\section{Ethical conduct}
For the proposed study, informed consent was obtained from all the subjects participating in the local acquisition to record their facial video and health data to be further used for research and development purposes in accordance with GDPR guidelines. The private dataset employed in the study was approved by the FotoNation ethics committee on 09/12/2021.

\section*{Acknowledgment}
The research conducted in this publication was funded by the Irish Research Council under project ID EBPPG/2021/92 as a part of the Employment-Based Programme Postgraduate Scholarship in partnership with FotoNation Ltd, Galway, Ireland. The research conducted in this publication was jointly funded by the Irish Research Council under grant number IRCLA/2023/1992.

\bibliographystyle{unsrt}  
\bibliography{references}
\end{document}